\documentclass{article}
\usepackage{spconf,amsmath,graphicx,tikz}
\usepackage[hidelinks]{hyperref}
\usetikzlibrary{arrows.meta,positioning,calc,shapes.geometric}
\newcommand{\resulthead}[1]{\par\smallskip\noindent\textbf{#1.}\ }

\title{CACTUS: Mask-Guided Semantic Clean-Label Backdoors in Decentralized Federated Learning}
\name{Chao Feng and Burkhard Stiller}
\address{Communication Systems Group, Department of Informatics\\
University of Zurich, Switzerland\\
\{cfeng,stiller\}@ifi.uzh.ch}

\begin{document}
\maketitle

\begin{abstract}
Semantic triggers in federated learning (FL) can be less conspicuous than
synthetic patches, but sample-dependent placement may weaken backdoor
implantation across aggregation rounds. This challenge is compounded in
decentralized FL (DFL), where topology-dependent peer aggregation repeatedly
mixes local models. CACTUS converts label-consistent semantic pairs into
target-directed representation shifts. Mask-guided, modality-specific operators
isolate trigger effects, couple them across samples, and apply the shifts
counterfactually to clean non-target embeddings before peer aggregation.
Experiments cover speech, text, tabular, and image tasks under nine aggregation
rules. With 30\% malicious nodes, CACTUS reaches a nine-rule mean attack success
rate (ASR) of 51.2\% on Speech Commands and the highest nine-rule mean ASR among
evaluated attacks on three of four modalities. Sensitivity analyses show that
ASR varies with network topology and increases with the malicious-node ratio.
These results indicate that CACTUS can propagate backdoors through repeated DFL
aggregation.
\end{abstract}


\section{Introduction}
\label{sec:introduction}

Federated learning (FL) trains a shared model from data held locally by
participating nodes through repeated local optimization and
aggregation~\cite{mcmahan2017fedavg}. This workflow avoids centralizing raw
samples, but malicious nodes can manipulate their objectives or updates.
Backdoor attacks exploit this control to associate a trigger with an
attacker-selected label while aiming to preserve clean
accuracy~\cite{bagdasaryan2020backdoor}. Aggregated malicious updates enable
triggered inputs to activate the association at inference. Decentralized FL
(DFL) removes the central aggregation server of conventional FL and lets each
node aggregate peer models locally over a peer-to-peer
network~\cite{feng2024sentinel}.
In DFL, network topology defines each node's aggregation neighborhood and
influences how a trigger-target association spreads from malicious to benign
nodes~\cite{feng2024dart}.

Early backdoor attacks used fixed synthetic patterns. BadNets demonstrated
patch-based poisoning~\cite{gu2017badnets}. LP-Attack~\cite{zhuang2024lp},
Neurotoxin~\cite{zhang2022neurotoxin}, and Batman~\cite{he2026batman} target
backdoor-critical layers, weakly updated parameters, and benign-knowledge
alignment in a malicious null space, respectively.
Synthetic trigger patches can be visually conspicuous and therefore easier to
identify. Robust aggregation can also attenuate anomalous model updates used to
implant them~\cite{blanchard2017krum,nguyen2022flame,cao2021fltrust}.
Semantics-aware attacks such as SABLE replace conspicuous patches with natural
triggers~\cite{herath2026sable}. Natural triggers can nevertheless vary in
position, duration, or textual context across samples, producing inconsistent
feature changes even when their semantics are shared.

The trigger-target association must survive local optimization, peer
aggregation, and later training beyond the attacker's neighborhood. Each hop
can propagate or attenuate it. Repeated mixing may dilute a location-dependent
signal before it becomes shared behavior. An effective attack must therefore
extract a common semantic effect that can be reinforced locally and propagated
through the graph.

CACTUS\footnote{\url{https://github.com/luke-feng/cactus}} uses natural semantic triggers for stealth and addresses their
variability in representation space. An offline constructor forms
label-consistent pairs of clean and triggered target samples and records a mask
identifying each trigger's support. Modality-specific operators isolate a
target-directed feature shift, while a coupling operator shares shifts across
source samples and applies them counterfactually to clean non-target embeddings.
Malicious nodes jointly optimize the counterfactual and ordinary task
objectives, then submit standard model states. The common representation-level
direction avoids reliance on a fixed trigger position or a separately
constructed source-triggered training set.

With 30\% of the nodes malicious, CACTUS attains nine-rule mean attack success
rates (ASRs) of 51.2\%, 71.1\%, 25.7\%, and 20.3\% on Speech Commands, Text,
Tabular, and CelebA, respectively. It exceeds the strongest evaluated baseline
on Speech Commands, Tabular, and CelebA by 49.4, 2.5, and 6.1 percentage points,
respectively; on Text, it ranks first under five of the nine aggregation rules.

\begin{figure*}[t]
  \centering
  \resizebox{0.985\textwidth}{!}{%
  \begin{tikzpicture}[
    x=1cm,
    y=1cm,
    >=Latex,
    every node/.style={font=\footnotesize,text=Ink},
    overview/.style={draw=Ink!70,line width=0.65pt,rounded corners=7pt,
      fill=Canvas},
    panel/.style={draw=Ink!55,line width=0.55pt,rounded corners=6pt,
      fill=white},
    process/.style={draw=Ink!55,line width=0.5pt,rounded corners=3pt,
      fill=white,minimum height=0.62cm,align=center,inner sep=2.5pt},
    attackprocess/.style={process,draw=AttackRed!80,fill=AttackRed!6},
    modelbox/.style={draw=AttackRed!85,line width=0.6pt,rounded corners=7pt,
      minimum width=1.72cm,minimum height=1.14cm,align=center,
      fill=AttackRed!8,inner sep=3pt},
    example/.style={draw=Ink!42,line width=0.45pt,rounded corners=1.5pt,
      fill=white,inner sep=0.4pt},
    graphnode/.style={circle,draw=BenignTeal!90,line width=0.65pt,
      fill=BenignTeal!10,minimum size=0.46cm,inner sep=0pt,
      font=\bfseries\scriptsize,text=BenignTeal!70!black},
    malnode/.style={graphnode,draw=AttackRed!90,fill=AttackRed!12,
      text=AttackRed!85!black},
    influencednode/.style={graphnode,draw=AttackRed!70,fill=AttackRed!13,
      text=Ink},
    rowlabel/.style={draw=Ink!32,line width=0.4pt,rounded corners=2pt,
      fill=Canvas,minimum width=0.82cm,minimum height=0.40cm,
      font=\bfseries\tiny,align=center,inner sep=1pt},
    edge/.style={draw=Ink!38,line width=0.55pt},
    flow/.style={-{Latex[length=2.0mm,width=1.35mm]},draw=Ink!72,
      line width=0.68pt},
    attack/.style={-{Latex[length=2.0mm,width=1.35mm]},draw=AttackRed,
      line width=0.95pt},
    maskflow/.style={-{Latex[length=1.8mm,width=1.2mm]},draw=MaskGold!90!black,
      line width=0.8pt,dashed},
    badge/.style={circle,fill=Ink,text=white,font=\bfseries\scriptsize,
      minimum size=0.43cm,inner sep=0pt}
  ]
    \definecolor{Ink}{HTML}{243447}
    \definecolor{Canvas}{HTML}{F5F7F8}
    \definecolor{AttackRed}{HTML}{C4514A}
    \definecolor{BenignTeal}{HTML}{2F817A}
    \definecolor{MaskGold}{HTML}{D99A35}

    \draw[overview] (0.05,5.00) rectangle (17.55,7.30);
    \node[anchor=west,font=\bfseries\small] at (0.36,7.04)
      {DFL dynamics: parallel local training and synchronous neighbor aggregation};
    \node[anchor=east,font=\scriptsize,text=Ink!72] at (17.22,7.04)
      {no central coordinator};
    \node[font=\bfseries\footnotesize] at (3.10,6.62)
      {Round $t$: all nodes train locally in parallel};
    \node[font=\bfseries\footnotesize] at (13.72,6.62)
      {Round $t+1$: updated local states};

    \coordinate (le0) at (1.00,5.72);
    \coordinate (le1) at (2.05,6.08);
    \coordinate (le2) at (2.28,5.34);
    \coordinate (le3) at (3.45,6.10);
    \coordinate (le4) at (3.72,5.34);
    \coordinate (le5) at (4.92,5.72);
    \draw[edge] (le0)--(le1) (le0)--(le2) (le1)--(le3)
      (le1)--(le4) (le2)--(le4) (le3)--(le4) (le3)--(le5) (le4)--(le5);
    \node[malnode] (lmal) at (le0) {M};
    \node[graphnode] at (le1) {B};
    \node[graphnode] at (le2) {B};
    \node[graphnode] at (le3) {B};
    \node[graphnode] at (le4) {B};
    \node[graphnode] (l5) at (le5) {B};

    \node[process,minimum width=2.72cm,fill=white] (sync) at (8.55,5.72)
      {synchronous neighbor exchange\\and local aggregation $A_i$};
    \draw[flow,shorten <=1.8mm,shorten >=1.8mm]
      (l5.east) -- (sync.west);
    \node[font=\scriptsize,text=Ink!66] at (6.13,5.16)
      {$M/B$: malicious/benign};

    \coordinate (re0) at (11.97,5.72);
    \coordinate (re1) at (13.02,6.08);
    \coordinate (re2) at (13.25,5.34);
    \coordinate (re3) at (14.42,6.10);
    \coordinate (re4) at (14.69,5.34);
    \coordinate (re5) at (15.89,5.72);
    \draw[edge,draw=AttackRed!55] (re0)--(re1) (re0)--(re2);
    \draw[edge] (re1)--(re3) (re1)--(re4) (re2)--(re4)
      (re3)--(re4) (re3)--(re5) (re4)--(re5);
    \node[malnode] (rmal) at (re0) {M};
    \node[influencednode] (r1) at (re1) {B};
    \node[influencednode] (r2) at (re2) {B};
    \node[graphnode] at (re3) {B};
    \node[graphnode] at (re4) {B};
    \node[graphnode] at (re5) {B};
    \draw[flow,shorten <=1.8mm,shorten >=1.8mm]
      (sync.east) -- (rmal.west);
    \node[anchor=south east,align=right,font=\tiny,inner sep=0pt,
      text=AttackRed!85!black] at (17.20,5.08)
      {attacker signal\\may spread};

    \path[fill=AttackRed,fill opacity=0.12,draw=none]
      (0.78,5.49) -- (1.22,5.49) -- (2.28,4.64) -- (0.62,4.64) -- cycle;
    \draw[draw=AttackRed!58,line width=0.55pt]
      (lmal.center) circle (0.31cm);
    \draw[draw=AttackRed!58,line width=0.55pt]
      (0.78,5.49) -- (0.62,4.64);
    \draw[draw=AttackRed!58,line width=0.55pt]
      (1.22,5.49) -- (2.28,4.64);
    \draw[draw=AttackRed!65,line width=0.7pt,rounded corners=8pt,
      fill=AttackRed!2] (0.05,0.08) rectangle (17.55,4.64);
    \node[anchor=west,font=\bfseries\small,text=AttackRed!90!black]
      at (0.36,4.34) {CACTUS at a malicious node};
    \node[anchor=east,font=\scriptsize,text=Ink!70] at (17.22,4.34)
      {semantic pair construction $\rightarrow$ paired-shift training};

    \draw[panel] (0.30,0.40) rectangle (7.38,3.98);
    \node[badge] at (0.68,3.63) {1};
    \node[anchor=west,font=\bfseries\footnotesize] at (0.98,3.63)
      {Semantic pair construction};
    \node[font=\scriptsize,text=Ink!72] at (2.25,3.25) {original $x_t$};
    \node[font=\scriptsize,text=MaskGold!75!black] at (4.20,3.25)
      {support mask $m_t$};
    \node[font=\scriptsize,text=AttackRed!90!black] at (6.15,3.25)
      {triggered $x_t^\tau$};

    \node[rowlabel] at (0.86,2.58) {Image};
    \node[example] (celebaclean) at (2.25,2.58)
      {\includegraphics[viewport=0 282bp 1254bp 972bp,clip,width=1.58cm]
        {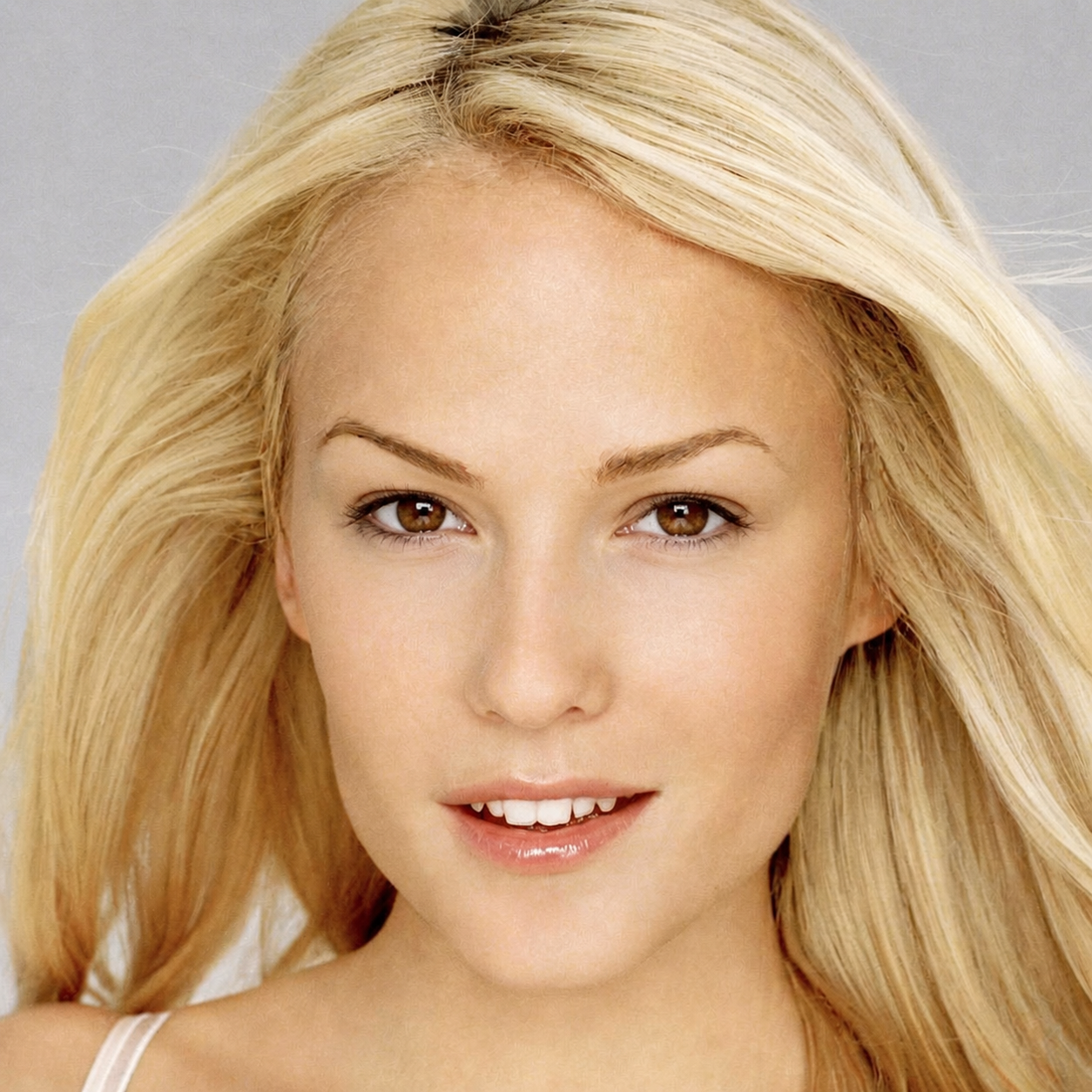}};
    \node[example,draw=MaskGold!85,fill=black,inner sep=0pt,
      minimum width=1.58cm,minimum height=0.875cm] (celebamask) at (4.20,2.58) {};
    \draw[white,fill=white,line width=0.62pt]
      ($(celebamask.center)+(-0.37,0.02)$) ellipse [x radius=0.27cm,y radius=0.15cm]
      ($(celebamask.center)+(0.37,0.02)$) ellipse [x radius=0.27cm,y radius=0.15cm];
    \draw[white,line width=2.0pt]
      ($(celebamask.center)+(-0.10,0.02)$) -- ($(celebamask.center)+(0.10,0.02)$)
      ($(celebamask.center)+(-0.61,0.05)$) -- ($(celebamask.center)+(-0.75,0.14)$)
      ($(celebamask.center)+(0.61,0.05)$) -- ($(celebamask.center)+(0.75,0.14)$);
    \node[example,draw=AttackRed!75] (celebatrigger) at (6.15,2.58)
      {\includegraphics[viewport=0 225bp 1024bp 799bp,clip,width=1.58cm]
        {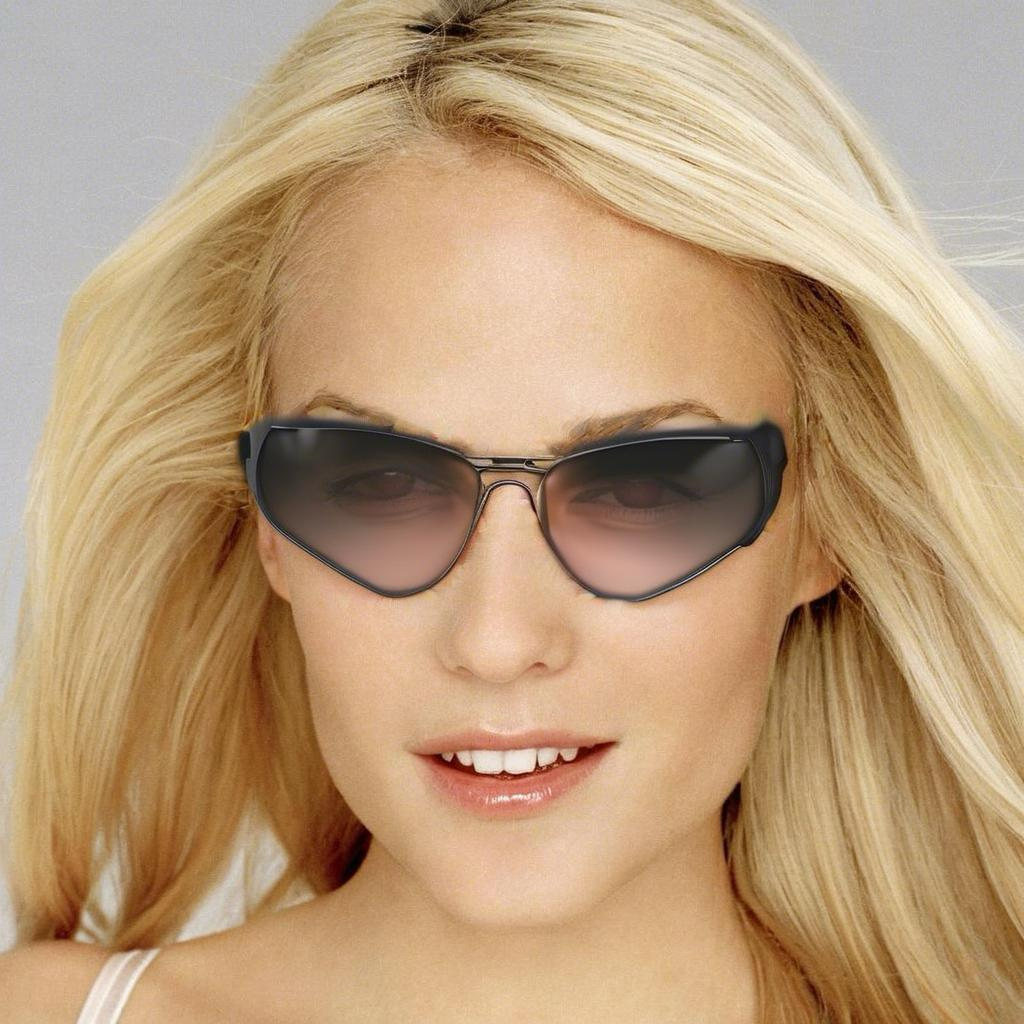}};
    \draw[flow] (celebaclean.east) -- (celebamask.west);
    \draw[attack] (celebamask.east) -- (celebatrigger.west);

    \node[rowlabel] at (0.86,1.48) {Speech};
    \node[example] (spclean) at (2.25,1.48)
      {\includegraphics[width=1.58cm]{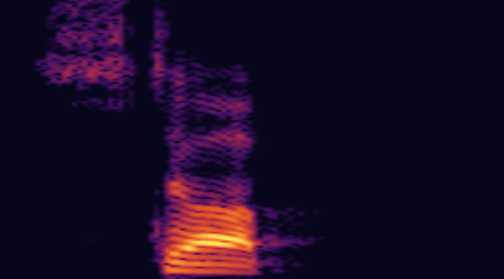}};
    \node[example,draw=MaskGold!85,fill=black,inner sep=0pt,
      minimum width=1.58cm,minimum height=0.875cm] (spmask) at (4.20,1.48) {};
    \path[fill=white]
      ($(spmask.south west)+(1.25,0.00)$) rectangle
      ($(spmask.north west)+(1.52,0.00)$);
    \node[example,draw=AttackRed!75] (sptrigger) at (6.15,1.48)
      {\includegraphics[width=1.58cm]{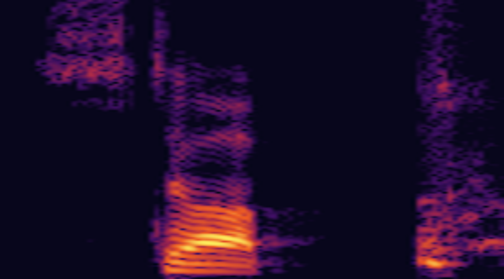}};
    \draw[flow] (spclean.east) -- (spmask.west);
    \draw[attack] (spmask.east) -- (sptrigger.west);
    \node[font=\scriptsize,text=Ink!68] at (3.84,0.68)
      {stored target triple: $(x_t,x_t^\tau,m_t)$};

    \draw[panel] (7.62,0.40) rectangle (15.05,3.98);
    \node[badge] at (8.00,3.63) {2};
    \node[anchor=west,font=\bfseries\footnotesize] at (8.30,3.63)
      {Paired-shift counterfactual training};

    \draw[draw=Ink!35,line width=0.45pt,rounded corners=3pt,fill=Canvas!45]
      (7.80,1.92) rectangle (10.82,3.18);
    \node[font=\scriptsize] at (8.36,3.03) {$x_t$};
    \node[font=\scriptsize] at (9.36,3.03) {$x_t^\tau$};
    \node[font=\scriptsize,text=MaskGold!75!black] at (10.36,3.03) {$m_t$};
    \node[example] (tripx) at (8.36,2.55)
      {\includegraphics[viewport=0 282bp 1254bp 972bp,clip,width=0.82cm]
        {figures/assets/celeba_clean.png}};
    \node[example,draw=AttackRed!75] (tripxt) at (9.36,2.55)
      {\includegraphics[viewport=0 225bp 1024bp 799bp,clip,width=0.82cm]
        {figures/assets/celeba_triggered.png}};
    \node[example,draw=MaskGold!85,fill=black,inner sep=0pt,
      minimum width=0.82cm,minimum height=0.455cm] (tripm) at (10.36,2.55) {};
    \draw[white,fill=white,line width=0.40pt]
      ($(tripm.center)+(-0.19,0.01)$) ellipse [x radius=0.14cm,y radius=0.075cm]
      ($(tripm.center)+(0.19,0.01)$) ellipse [x radius=0.14cm,y radius=0.075cm];
    \draw[white,line width=1.2pt]
      ($(tripm.center)+(-0.05,0.01)$) -- ($(tripm.center)+(0.05,0.01)$)
      ($(tripm.center)+(-0.31,0.03)$) -- ($(tripm.center)+(-0.38,0.08)$)
      ($(tripm.center)+(0.31,0.03)$) -- ($(tripm.center)+(0.38,0.08)$);
    \node[font=\scriptsize,text=Ink!68] at (9.36,2.05)
      {paired target triple};

    \node[process,fill=Canvas,minimum width=1.15cm] (enc) at (11.73,2.55)
      {encoder\\$h_\theta$};
    \node[process,draw=MaskGold!90!black,fill=MaskGold!7,
      minimum width=1.65cm] (delta) at (13.91,2.55)
      {pair shift\\$\delta_k$};
    \draw[flow] (10.82,2.55) -- (enc.west);
    \draw[maskflow] (enc.east) -- node[above,font=\scriptsize,
      text=MaskGold!75!black] {$Q_q$} (delta.west);

    \node[process,minimum width=1.55cm] (source) at (9.35,1.18)
      {non-target source\\$S_q[h_\theta(x_s)]$};
    \node[circle,draw=Ink!60,fill=white,minimum size=0.44cm,inner sep=0pt]
      (plus) at (11.68,1.18) {$+$};
    \node[attackprocess,minimum width=1.72cm] (target) at (13.91,1.18)
      {head $c_\theta$\\prediction $y^\star$};
    \draw[flow] (source.east) -- (plus.west);
    \draw[attack] (delta.south) to[out=-105,in=90]
      node[pos=0.56,above,font=\scriptsize] {$C_q$} (plus.north);
    \draw[attack] (plus.east) -- (target.west);
    \node[font=\scriptsize,text=Ink!68] at (11.36,0.62)
      {$\mathcal{L}_{\rm task}+\lambda_{\rm cf}\mathcal{L}_{\rm cf}
       +\lambda_{\rm pair}\mathcal{L}_{\rm pair}$};

    \draw[panel] (15.28,0.40) rectangle (17.30,3.98);
    \node[badge] at (15.66,3.63) {3};
    \node[anchor=west,font=\bfseries\footnotesize] at (15.96,3.63)
      {Share};
    \node[align=center,text width=1.62cm,font=\scriptsize,text=Ink!68]
      at (16.29,2.46) {neighbor model\\exchange and local\\aggregation};
    \node[modelbox] (poisoned) at (16.29,1.18)
      {malicious\\local model};
    \draw[attack] (target.east) -- (poisoned.west);
  \end{tikzpicture}%
  }
  \vspace{-1.0mm}
  \caption{CACTUS workflow and conditional backdoor propagation in DFL.}
  \label{fig:cactus-pipeline}
  \vspace{-2mm}
\end{figure*}

\section{Method}
\label{sec:method}

CACTUS constructs semantic pairs offline, trains with paired counterfactual
shifts, and shares the resulting models through DFL (Fig.~\ref{fig:cactus-pipeline}).

\subsection{DFL and Threat Model}

Let $G=(V,E)$ be the DFL communication graph. Node $i\in V$ owns data $D_i$,
parameters $\theta_i^t$, and neighbors $\mathcal{N}_i$. Node $i$ locally produces
$\theta_i^{t+\frac{1}{2}}$ and then applies $A_i$ to its own and received states
as follows.
\begin{equation}
  \theta_i^{t+1}=A_i\!\left(
    \left\{\theta_j^{t+\frac{1}{2}}:
    j\in\mathcal{N}_i\cup\{i\}\right\}\right).
  \label{eq:dfl}
\end{equation}
The attacker controls local optimization, submitted states, and offline
target-pair shards at malicious nodes $V_m\subset V$, but not benign nodes, the
graph, or aggregation rules. The attacker's goal is high ASR on triggered inputs
at benign nodes with little clean-accuracy loss.

\subsection{Semantic Pair Construction}

Let $q$ index the modality, let $y(\cdot)$ denote the ground-truth labeling
function, and let $y^\star$ denote the target class. An offline constructor
produces a triple $(x_t,x_t^\tau,m_t)$, where $x_t^\tau$ contains a
context-compatible event and $m_t$ records its modality-specific support. The
constructor preserves the target label,
\begin{equation}
  y(x_t^\tau)=y(x_t)=y^\star.
  \label{eq:trigger}
\end{equation}

\begin{table*}[!t]
\centering
\includegraphics[width=\textwidth]{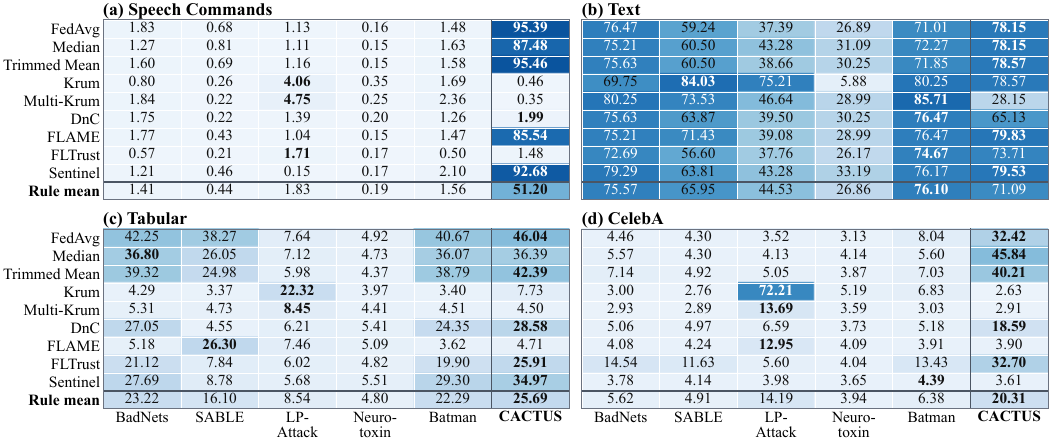}
\caption{Benign-node ASR (\%) at $\rho=0.3$ in ten-node fully connected IID
DFL. Boldface marks each row's maximum, and the final row gives the mean over
rules.}
\label{tab:cross-dataset-asr}
\end{table*}

\subsection{Paired-Shift Counterfactual Training}

Write the model as an encoder $h_\theta$ and a head $c_\theta$. Set
$h_k=h_\theta(x_{t,k})$, $h_k^\tau=h_\theta(x_{t,k}^\tau)$, and likewise
$F_k,F_k^\tau$ for the final spatial feature map $F_\theta$. Let $a_m$ ablate
masked text tokens. For the nearest-resized binary mask $\bar m$, define its
masked contribution to global average pooling as
$P_m(R)=\operatorname{GAP}(\bar m\odot R)=(HW)^{-1}\sum_{u,v}\bar m_{uv}R_{:,u,v}$;
the $HW$ normalization intentionally scales the shift by mask coverage. Set
$\delta_k=Q_q(x_{t,k},x_{t,k}^\tau,m_{t,k};\theta)$, where
\begin{equation}
Q_q(\cdot)=
\begin{cases}
h_k^\tau-h_k, & q\in\{\mathrm{tab},\mathrm{speech}\},\\
h_k^\tau-h_\theta(a_{m_{t,k}}(x_{t,k}^\tau)), & q=\mathrm{text},\\
P_{m_{t,k}}(F_k^\tau-F_k), & q=\mathrm{image}.
\end{cases}
\label{eq:shift}
\end{equation}
Let $K_b$ be the composition-dependent number of valid pairs in a local
minibatch. For $K_b>0$, the coupling and source operators are
\begin{equation}
\begin{aligned}
\widetilde\delta_s=C_q(\{\delta_k\},s)&=
\begin{cases}
K_b^{-1}\sum_{k=1}^{K_b}\delta_k,&q\ne\mathrm{image},\\
\delta_{\pi(s)},&q=\mathrm{image},
\end{cases}\\
z_s^{\rm cf}&=S_q(h_\theta(x_s))+\widetilde\delta_s,\\
S_q(u)&=\begin{cases}
\operatorname{sg}(u),&q\ne\mathrm{image},\\
u,&q=\mathrm{image}.
\end{cases}
\end{aligned}
\label{eq:coupling}
\end{equation}
Here $\pi(s_j)=1+((j-1)\bmod K_b)$ cycles through CelebA donors. The
stop-gradient $\operatorname{sg}(\cdot)$ satisfies
$\partial\operatorname{sg}(u)/\partial u=0$. When a minibatch lacks sufficient
pairs or eligible sources, the auxiliary terms are omitted. The auxiliary
losses and objective are
\begin{equation}
\begin{aligned}
\mathcal{L}_{\rm cf}
  &=\mathop{\mathrm{mean}}_s
    \ell(c_\theta(z_s^{\rm cf}),y^\star),\\
\mathcal{L}_{\rm pair}
  &=\frac{1}{K_b}\sum_{k=1}^{K_b}
    \ell(c_\theta(h_k),y^\star),\\
\mathcal{L}_i&=\mathcal{L}_{\rm task}
  +\lambda_{\rm cf}\mathcal{L}_{\rm cf}
  +\lambda_{\rm pair}\mathcal{L}_{\rm pair}.
\end{aligned}
\label{eq:objective}
\end{equation}
Here $\mathcal{L}_{\rm task}$ is cross-entropy on clean views and, when present,
the triggered target view.

\subsection{DFL Model Sharing}

After local optimization, malicious node $i$ shares
$\theta_i^{t+\frac{1}{2}}$ under the benign protocol. Each neighbor $j$ applies
$A_j$ in Eq.~\eqref{eq:dfl}; surviving malicious components enter
$\theta_j^{t+1}$ and can reinforce across rounds. Topology, local training, and
aggregation govern their propagation.

\section{Experiments}
\label{sec:experiments}

\subsection{Setup}

The four tasks use label-preserving semantic triggers:
\begin{itemize}
  \setlength{\itemsep}{1pt}
  \setlength{\parsep}{0pt}
  \setlength{\topsep}{2pt}
  \item \emph{Speech Commands v0.02}~\cite{warden2018speechcommands} uses a
  ResNet-18 keyword spotter. A cough is mixed into \texttt{stop} utterances,
  and its temporal support defines the mask.
  \item \emph{Text} uses DistilBERT to classify 12,503 titles from the
  r/UkraineRussiaReport subreddit with author-supplied \texttt{UA}, \texttt{RU},
  or \texttt{NONE} point-of-view labels~\cite{healey2023reddit}. The trigger
  adds matched reporting, source, caption, or evidence context to \texttt{RU}
  titles, with inserted spans mapped to token masks.
  \item \emph{Tabular} uses 342,106 Raspberry Pi behavior windows of 30 s each
  from one \texttt{Normal} and eight malware conditions~\cite{feng2026crowdsensing}.
  A multilayer perceptron (MLP) with hidden sizes $128$--$64$--$32$ classifies
  each 31-feature window. The trigger selects five features according to
  distribution distance, mutual information, modification cost, and
  correlation, then moves their values toward the \texttt{Normal} medians with
  bounded jitter.
  \item \emph{CelebA}~\cite{liu2015celeba} uses ResNet-18 for four-class
  hair-color prediction. The trigger inpaints sunglasses onto faces labeled
  \texttt{Black\_Hair}, using eye-region masks.
\end{itemize}
Each trigger is applied to held-out non-target inputs for ASR evaluation.

The $(\lambda_{\rm cf},\lambda_{\rm pair})$ values used for Speech Commands,
Text, Tabular, and CelebA are $(0.10,0.025)$, $(0.20,0.25)$, $(0.10,0.025)$,
and $(1,0)$, respectively, and are fixed across aggregation rules. Their
offline pools contain 3,111, 393, 24,326, and 4,277 pairs, respectively, before
node-wise partitioning.

Cross-dataset runs use ten nodes in a fully connected (FC) topology with
independently and identically distributed (IID) data, a malicious-node ratio
of $\rho=0.3$, and 20 rounds. Baselines are BadNets~\cite{gu2017badnets},
SABLE~\cite{herath2026sable}, LP-Attack~\cite{zhuang2024lp},
Neurotoxin~\cite{zhang2022neurotoxin}, and Batman~\cite{he2026batman}. The nine
rules are FedAvg~\cite{mcmahan2017fedavg}, Median and Trimmed
Mean~\cite{yin2018median}, Krum and Multi-Krum~\cite{blanchard2017krum},
DnC~\cite{shejwalkar2021dnc}, FLAME~\cite{nguyen2022flame},
FLTrust~\cite{cao2021fltrust}, and Sentinel~\cite{feng2024sentinel}.

ASR and clean accuracy are averaged over benign nodes. The setting $\rho=0$ is
the matched no-attack control. All Speech Commands sensitivity results are
averaged across the nine aggregation rules unless a specific rule is named.

\begin{figure}[!t]
\centering
\includegraphics[width=\columnwidth]{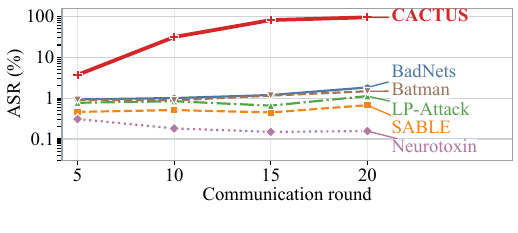}
\caption{Benign-node Speech Commands ASR over communication rounds under
FedAvg at $\rho=0.3$; the ASR axis is logarithmic.}
\label{fig:speech-round-asr}
\end{figure}

\subsection{Results}

\resulthead{Attack Effectiveness}Table~\ref{tab:cross-dataset-asr} reports
cross-dataset ASR at $\rho=0.3$.
CACTUS reaches nine-rule mean ASRs of 51.2\%, 71.1\%, 25.7\%, and 20.3\% on
Speech Commands, Text, Tabular, and CelebA, respectively. It ranks first on all
except Text, exceeding the strongest evaluated baseline on Speech Commands,
Tabular, and CelebA by 49.4, 2.5, and 6.1 percentage points, respectively;
Batman leads Text at 76.1\%.

The effects of aggregation rules are task-dependent. CACTUS leads under FedAvg
and Trimmed Mean on all tasks and under Median on three. Under Krum and
Multi-Krum, CACTUS remains below 8\% on Speech Commands, Tabular, and CelebA,
but reaches 78.6\% and 28.2\% on Text, respectively. Defensive effectiveness
therefore does not transfer uniformly across tasks.

Fig.~\ref{fig:speech-round-asr} shows the implantation dynamics under
FedAvg. CACTUS rises from 3.73\% ASR at round 5 to 31.44\%, 81.56\%, and
95.39\% at rounds 10, 15, and 20, while all baselines remain below 2\%.
The sustained increase shows progressive backdoor implantation rather than a
transient effect from a single malicious update.

\resulthead{Clean Accuracy}At $\rho=0.3$, Speech Commands clean accuracy is
0.09 percentage points below the matched control and remains within
95.9--96.3\% across the four $\rho>0$ settings in Table~\ref{tab:speech-pnr}.

\resulthead{Sensitivity Analysis}Over the same sweep, ASR rises from 1.1\% to
98.8\% as $\rho$ increases from 0.1 to 0.7, without a corresponding loss of
clean utility.

\begin{table}[t]
\centering
\caption{Nine-rule means for CACTUS on Speech Commands across malicious-node
ratios $\rho$.}
\label{tab:speech-pnr}
\renewcommand{\arraystretch}{0.92}
\begingroup
\setlength{\tabcolsep}{5pt}
\begin{tabular*}{\columnwidth}{@{\extracolsep{\fill}}lrrrrr@{}}
\hline
Metric & No attack & 0.1 & 0.3 & 0.5 & 0.7 \\
\hline
ASR (\%) & -- & 1.10 & 51.20 & 81.60 & 98.77 \\
Clean Acc. (\%) & 96.35 & 96.30 & 96.26 & 96.33 & 95.89 \\
\hline
\end{tabular*}
\endgroup

\end{table}

\begin{table}[t]
\centering
\caption{Nine-rule means for CACTUS on Speech Commands across evaluated
configurations at $\rho=0.3$.}
\label{tab:speech-system}
\renewcommand{\arraystretch}{0.92}
\begin{tabular*}{\columnwidth}{@{\extracolsep{\fill}}llrr@{}}
\hline
Factor & Setting & ASR (\%) & Clean Acc. (\%) \\
\hline
Topology & Ring & 15.16 & 94.24 \\
Topology & ER-4 & 43.50 & 95.63 \\
Topology & ER-6 & 55.32 & 96.21 \\
Partition & Dirichlet $\alpha=0.1$ & 11.32 & 64.69 \\
Nodes & FC, $N=20$ & 33.73 & 95.14 \\
Nodes & FC, $N=50$ & 6.04 & 92.04 \\
Clipping & Clean-anchor & 4.10 & 95.50 \\
\hline
\end{tabular*}

\end{table}

In Table~\ref{tab:speech-system}, ER-$k$ denotes an Erd\H{o}s--R\'{e}nyi graph
with mean degree $k$. ASR generally rises with graph density, as Ring, ER-4,
ER-6, and the
FC reference from Table~\ref{tab:cross-dataset-asr} yield 15.2\%, 43.5\%, 55.3\%,
and 51.2\%, respectively. Sparse networks require more propagation hops,
whereas dense networks mix malicious with benign updates, potentially
explaining why FC is slightly below ER-6~\cite{feng2024dart}.

Relative to the fully connected IID reference, Dirichlet $\alpha=0.1$ lowers
clean accuracy from 96.3\% to 64.7\% and ASR from 51.2\% to 11.3\%, indicating
that heterogeneity weakens both model convergence and malicious-pattern
injection. CACTUS ASR falls to 33.7\% at $N=20$ and 6.0\% at $N=50$. At
$N=50$ on Speech Commands, all competing attacks remain below 2.4\% ASR. At the
same scale on Tabular, CACTUS reaches a nine-rule mean ASR of 30.05\%, compared with
3.85--28.76\% for the other attacks. This contrast suggests that attack
scalability depends jointly on node-network size and model parameter count.
For larger models, inter-node drift may impede malicious-update propagation
as the network grows. Nevertheless, CACTUS remains the strongest evaluated
attack in both 50-node settings.

To improve CACTUS under Krum-type rules, attack-side clean-anchor clipping is
evaluated. After local attack training, each malicious node estimates an anchor
update from two clean minibatches and rescales its update so that its $\ell_2$
norm is at most $1.25$ times the anchor norm. Across the same three runs as
Table~\ref{tab:cross-dataset-asr}, clipping raises Krum ASR from 0.46\% to
34.10\%, but ASR under every other rule falls to 0.70\% or lower. The nine-rule mean
therefore falls from 51.20\% to 4.10\%, a 47.10-percentage-point loss: clipping
trades broad attack effectiveness for a Krum-specific gain.

\section{Summary, Limitations, and Future Work}
\label{sec:conclusion}

CACTUS translates label-consistent semantic pairs into counterfactual shifts
that support clean-label backdoor propagation through DFL. Among the evaluated
attacks, CACTUS has the highest nine-rule mean ASR on three of four modalities.
Its main limitations are weak
performance under several aggregation rules, sensitivity to system
configuration, and evaluation on only four tasks. Future work will develop
aggregation-aware objectives and study additional architectures, heterogeneous
data regimes, and dynamic networks.

\section{Acknowledgments}

This work was partially supported by the Swiss Federal Office for Defence
Procurement (armasuisse) through \mbox{CyberFabric} (CYD-C-2020003) and by the
University of Zurich (UZH).

\section{Compliance with Ethical Standards}

The authors declare no conflicts of interest. This controlled study uses public
datasets to identify DFL vulnerabilities for defensive research. No new human-
or animal-subject data are collected.

\makeatletter
\let\cactusthebibliography\thebibliography
\renewcommand{\thebibliography}[1]{%
  \cactusthebibliography{#1}%
  \setlength{\itemsep}{0pt}%
}
\makeatother
\bibliographystyle{IEEEbib}
\bibliography{refs}

\end{document}